\pdfoutput=1
\documentclass[11pt]{article}

\usepackage[preprint]{acl}

\usepackage{times}
\usepackage{latexsym}

\usepackage[T1]{fontenc}
\usepackage[utf8]{inputenc}

\usepackage{microtype}

\usepackage{inconsolata}

\usepackage{graphicx}

\usepackage{fontawesome5}

\usepackage{booktabs}
\usepackage{multirow}
\usepackage{float}
\usepackage[normalem]{ulem}
\useunder{\uline}{\ul}{}

\usepackage{stfloats}

\title{Bemba Speech Translation: Exploring a Low-Resource African Language}

\author{
  \hspace{2em}Muhammad Hazim Al Farouq \\
  \hspace{2em}{\small Kreasof AI} \\
  \hspace{2em}{\small Research Labs} \\
  \hspace{2em}{\small Jakarta, Indonesia} \\
  \\\And
  \hspace{3.5em}Aman Kassahun Wassie \\
  \hspace{3.5em}{\small African Institute for} \\
  \hspace{3.5em}{\small Mathematical Sciences (AIMS)} \\
  \hspace{3.5em}{\small Addis Ababa, Ethiopia} \\
  \\\And
  \hspace{0.5em}Yasmin Moslem\textsuperscript{\tiny\faStar[regular]} \\
  {\small ADAPT Centre} \\
  {\small Trinity College Dublin} \\
  {\small Dublin, Ireland} \\
  \\
  }

\begin{document}
\maketitle

\def\thefootnote{{\scalebox{0.5}{\faStar[regular]}}}
\footnotetext{Correspondence: \url{yasmin [at] machinetranslation.io}}\def\thefootnote{\arabic{footnote}}

\begin{abstract}
This paper describes our system submission to the International Conference on Spoken Language Translation (IWSLT 2025), low-resource languages track, namely for Bemba-to-English speech translation. We built cascaded speech translation systems based on Whisper and NLLB-200, and employed data augmentation techniques, such as back-translation. We investigate the effect of using synthetic data and discuss our experimental setup. 
\end{abstract}

\section{Introduction}

Low-resource languages face critical limitations due to the scarcity and scattered nature of the available data \citep{Haddow2022-LowResourceMT}. Speech translation for low-resource languages involves similar challenges \citep{Ahmad2024-IWSLT,Moslem2024-IWSLT,SeaCrowd2024,abdulmumin-etal-2025-findings}, Similarly, speech applications for African languages are very limited due to the lack of linguistic resources. For example, Bemba is an under-resourced language spoken by over 30\% of the population in Zambia \citep{Sikasote-2023-BembaSpeech}. Hence, the IWSLT shared task on speech translation for low-resource languages aims to benchmark and promote speech translation technology for a diverse range of dialects and low-resource languages.

We participated in the Bemba-to-English language pair through building cascaded speech translation systems. In other words, we employed Whisper \citep{Radford2022-Whisper} for automatic speech recognition (ASR), and NLLB-200 \citep{NLLB2022} for text-to-text machine translation (MT). For ASR, we fine-tuned Whisper models using two datasets, BembaSpeech and BIG-C. For MT, we fine-tuned the NLLB-200 models using the bilingual segments of the BIG-C dataset, and the “dev” split of the FLORES-200 dataset. In addition, we augmented the Bemba-to-English training data with back-translation of a portion of the Tatoeba dataset from English into Bemba. The back-translated data was filtered based on cross-entropy scores. As Table \ref{tab:eval-submissions} shows, the systems we submitted to the shared tasks are as follows:
\begin{itemize}
    \item Primary: It uses Whisper-Medium for ASR and NLLB-200 3.3B for MT.
    \item Contrastive 1: It uses Whisper-Small for ASR and NLLB-200 3.3B for MT.
    \item Contrastive 2: It uses Whisper-Small for ASR and NLLB-200 600M for MT.
\end{itemize}

\section{Data}

The data we used to train our Bemba-to-English speech translation models can be categorized into: (1) authentic data, and (2) synthetic data. The following sections provide more details (cf. Table~\ref{tab:data}).

\setlength{\tabcolsep}{3.5pt} % Default value: 6pt

\def\ci#1{\textcircled{\resizebox{.5em}{!}{#1}}}
\def\cis#1{\textcircled{\resizebox{.4em}{!}{#1}}}
\def\ciCheck{\ci{\faCheck}}
\def\ciTimes{\cis{\faTimes}}

\begin{table}[H]
\centering
\begin{small}
\begin{tabular}{@{}llrrrccc@{}}
\toprule
\textbf{Dataset} & \textbf{Language} & \textbf{Train} & \textbf{Dev} & \textbf{Test} & \textbf{Audio} \\ \midrule
\textbf{Big-C }      & Bem-Eng  & 82,371 & 2,782 & 2,763  & \ciCheck   \\
\textbf{BembaSpeech} & Bem      & 12,421 & 1,700 & 1,359  & \ciCheck  \\
\textbf{FLORES-200}  & Bem-Eng  & 997    & 0     & 1,012  & \ciTimes  \\
\textbf{Tatoeba}     & Eng      & 20,121 & 0     & 0      & \ciTimes  \\ \bottomrule
\end{tabular}
\end{small}
\caption{Data Statistics: The “Language” column specifies which languages are originally available in each dataset. “Train”, “Dev”, and “Test” represent the dataset sizes. The “Audio” column indicates whether each dataset includes audio signals.}
\label{tab:data}
\end{table}
\begin{table*}[htp]
\centering
\begin{tabular}{@{}lrrrrrr@{}}
\toprule

  &
  \multicolumn{3}{c}{\textbf{FLORES-200}} &
  \multicolumn{3}{c}{\textbf{BIG-C}} \\ 
  \cmidrule(rl){2-4} \cmidrule(rl){5-7} \addlinespace[3pt]

\textbf{Training Dataset(s)} & \textbf{BLEU} & \textbf{chrF++} & \textbf{COMET}
                             & \textbf{BLEU} & \textbf{chrF++} & \textbf{COMET} \\ \midrule
Big-C                        & 18.13 & 42.11 & 53.25 
                             & \underline{27.83} & \underline{51.08} & 53.28 \\
Big-C + Tatoeba             & \underline{21.67} & \underline{45.25} & \underline{55.64} & 27.82 & 50.98 & \textbf{53.39} \\ \midrule
Big-C + FLORES-200           & 25.21 & 47.31 & 57.23
                             & 27.96 & 51.03 & \underline{53.29} \\
Big-C + FLORES-200 + Tatoeba & \textbf{25.70} & \textbf{47.75} & \textbf{58.29}
                             & \textbf{28.60} & \textbf{51.38} & 53.08 \\ \bottomrule
\end{tabular}
\caption{MT Evaluation: In general, the models trained with both authentic data (Big-C \& FLORES-200) and back-translated data (Tatoeba) outperform the models trained with the authentic data only. All the models in this table uses NLLB-200 600M.}
\label{tab:eval-mt}
\end{table*}

% All of these models are (without <bt>tag) and they use filtered back-translated data.

\subsection{Authentic Data}
\label{sec:authentic}

We filtered the authentic data by removing any overlaps between the training data and test data based on the text transcript. For building our models, we used the following data sources.

\begin{itemize}
    \item \textbf{Big-C} is a parallel corpus of speech and transcriptions of image-grounded dialogues between Bemba speakers and their corresponding English translations. It contains 92,117 spoken utterances of both complete and incomplete dialogues, amounting to 187 hours of speech data grounded on 16,229 unique images. The dataset aims to enable the development of speech recognition, speech, and text translation systems for Bemba, as well as facilitate research in language grounding and multimodal model development \citep{Sikasote-2023-BembaSpeech}.\footnote{\url{https://github.com/csikasote/bigc}} Since this dataset includes audio and transcription in Bemba as well as translation into English, we could use it to build both modules of our cascaded systems, i.e. ASR and MT. Table \ref{tab:examples} shows examples of sentence pairs from the Big-C datasets.
    
    \item \textbf{BembaSpeech} is an ASR corpus for the Bemba language of Zambia. It contains read speech from diverse publicly available Bemba sources; literature books, radio/TV shows transcripts, YouTube video transcripts as well as various open online sources. Its purpose is to enable the training and testing of automatic speech recognition (ASR) systems in Bemba language. The corpus has 14,438 utterances, culminating into 24.5 hours of speech data \citep{Sikasote-2023-BigC}.\footnote{\url{https://github.com/csikasote/BembaSpeech}} We used the BembaSpeech dataset in addition to the Big-C dataset to build our ASR models.
    
    \item \textbf{FLORES-200} \citep{Goyal2022-Flores-spBLEU} is a bilingual text-only dataset for machine translation. We used the Bemba-to-English “dev” split for training, and the “devtest” split for testing. 
    
    \item \textbf{Tatoeba} \citep{Tiedemann2020-Tatoeba} is a monolingual dataset in English. We used a portion of it for back-translation (cf. Section \ref{sec:synthetic}).
\end{itemize}

\subsection{Synthetic Data}
\label{sec:synthetic}

We augmented our authentic data (cf. Section \ref{sec:authentic}) with synthetic data created with back-translation. To this end, we fine-tuned the NLLB-200 600M model in the other direction, i.e. for the English-to-Bemba language pair. Thereafter, we translated the English sentences from Tatoeba into Bemba using the fine-tuned English-to-Bemba NLLB-200 model. For translation, we used CTranslate2 \citep{Klein2020-Efficient}, generating the prediction cross-entropy scores for each sentence, and calculating the exponential of the scores for better readability. We filtered data based on the cross-entropy scores, removing low-quality segments. We removed segments with scores less than 0.77 based on manual exploration of samples of the generated back-translations. While the unfiltered back-translated data consists of 85,000 segments, the filtered back-translated data consists of 20,000 segments. Finally, we prepended the source side (Bemba) with the \textit{<bt>} tag to indicate that the data is synthetic. Moreover, we experimented with removing the \textit{<bt>} tag and found that this achieves slightly better results when testing with the FLORES-200's “devtest” split, as the data was already filtered (cf. Table \ref{tab:eval-bt}).

\def\ci#1{\textcircled{\resizebox{.5em}{!}{#1}}}
\def\cis#1{\textcircled{\resizebox{.4em}{!}{#1}}}
\def\ciCheck{\ci{\faCheck}}
\def\ciTimes{\cis{\faTimes}}

\begin{table*}[htbp]
\centering

\resizebox{\textwidth}{!}{
\begin{tabular}{@{}crccrrrrrr@{}}
\toprule

  &
  & 
  &
  &
  \multicolumn{3}{c}{\textbf{FLORES-200}} &
  \multicolumn{3}{c}{\textbf{BIG-C}} \\ 
  \cmidrule(rl){5-7} \cmidrule(rl){8-10} \addlinespace[3pt]

  \textbf{Datasets} &
  \textbf{BT Size} &
  \textbf{Filtered} &
  \textbf{<bt> tag} &
  \textbf{BLEU} &
  \textbf{chrF++} &
  \textbf{COMET} &
  \textbf{BLEU} &
  \textbf{chrF++} &
  \textbf{COMET} \\ \midrule
  \multirow{3}{*}{BIG-C + Tatoeba} 
                            & 85,155 & \ciTimes & \ciCheck & 20.96  & 45.06 & \textbf{55.92} & \textbf{28.17} & \textbf{51.26} & 53.45 \\
                            & 20,121 & \ciCheck & \ciCheck & 19.82 & 44.09 & 54.79 
                            & 28.04 & 51.20 & \textbf{53.51} \\
                            & 20,121 & \ciCheck & \ciTimes & \textbf{21.67} & \textbf{45.25} & \underline{55.64} & 27.82 & 50.98 & 53.39  \\ \bottomrule
\end{tabular}
}
\caption{Performance of MT models that are based on NLLB-200 600M and trained using both authentic data and augmented back-translated data. There are two pre-processing aspects applied to the augmented data, filtering the data based on cross-entropy scores, and prepending the source sentence with the \textit{<bt>} tag. Evaluating the models with the devtest split of the FLORES-200 dataset, the highest evaluation scores, in terms BLEU and chrF++, are achieved when the back-translated data is filtered and the \textit{<bt>} tag is removed. Meanwhile, the AfriCOMET score (COMET) of this model is comparable to the model where the back-translated data is not filtered and the source is prepended with the \textit{<bt>} tag. Evaluating the models with the hold-out test split of Big-C reveals a different outcome where using the \textit{<bt>} tag results in relatively higher scores, although the scores of the three experiments are relatively comparable.
It is worth noting that the filtered back-translated data consists of only 20k segments, while the unfiltered back-translated data consists of 85k segments.}
\label{tab:eval-bt}
\end{table*}

%It is worth noting though that AfriCOMET does not officially support the Bemba language.

\section{Experiments and Results}

\begin{figure}[thp]
\centering
  \includegraphics[width=0.9\linewidth]{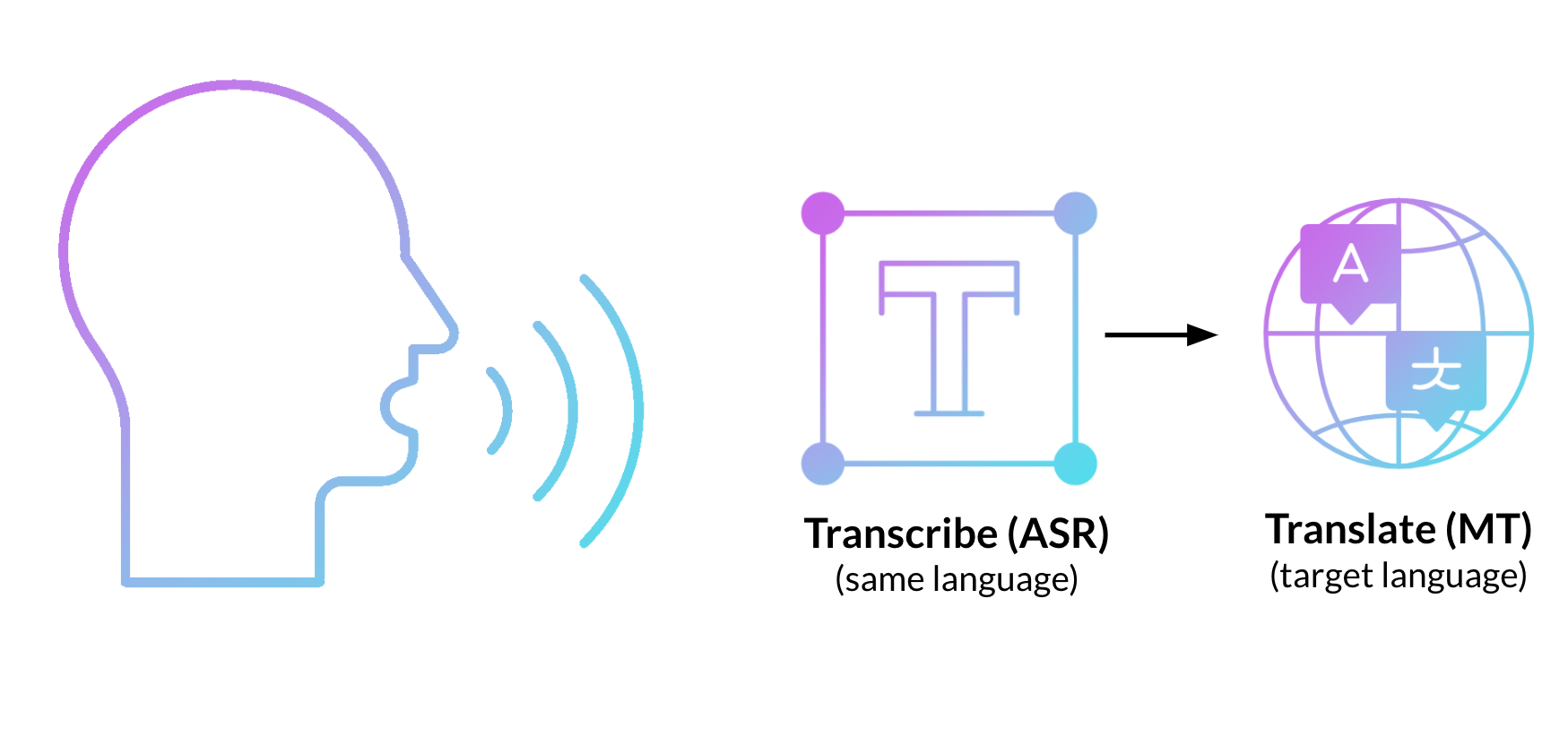}
  \caption{Cascaded speech translation systems use two models, an ASR model to generate audio transcriptions in the same language, and then an MT model to translate the generated transcriptions into the target language.}
  \label{fig:cascaded}
\end{figure}

As illustrated by Figure \ref{fig:cascaded}, our cascaded systems involve two components, an ASR model based on Whisper to generate transcriptions and an MT model based on NLLB-200 to generate text translation. We experimented with different versions of these models, namely Whisper Small and Medium, and NLLB-200 with 600M and 3.3B parameters. Our code for data preparation, training, and evaluation is publicly available.\footnote{\url{https://github.com/cobrayyxx/Bemba-IWSLT2025}}

\paragraph{Training:} We trained our models for 3 epochs, saving the best checkpoint based on the chrF++ score during training on the validation dataset. Our training arguments were chosen based on both manual exploration and automatic hyperparameter optimization using the Optuna framework \citep{Akiba-2019-Optuna}. The most important arguments are a learning rate of 1e-4 and a warm-up ratio of 0.03.

\paragraph{Inference:} For inference, we used Faster-Whisper \footnote{\url{https://github.com/SYSTRAN/faster-whisper}} with the default VAD\footnote{Voice Audio Detection (VAD) removes low-amplitude samples
from an audio signal, which might represent silence or noise.} arguments, and 5 for the ``beam size''. The model was quantized with the float16 precision for more efficient inference.

\paragraph{Evaluation:} To evaluate our systems, we calculated BLEU \citep{Papineni2002-BLEU},  and chrF++ \citep{Popovic2017-chrF++}, as implemented in the sacreBLEU library\footnote{\url{https://github.com/mjpost/sacrebleu}} \citep{Post2018-sacreBLEU}. For semantic evaluation, we used AfriCOMET \citep{Wang2024-AfriCOMET}. We conducted ASR evaluation (cf. Table~\ref{tab:eval-asr}) and MT evaluations (cf. Table~\ref{tab:eval-mt} and Table~\ref{tab:eval-bt}). Finally, we evaluated the whole cascaded systems (cf. Table~\ref{tab:eval-submissions}).

% Please add the following required packages to your document preamble:
% \usepackage{booktabs}
% \begin{table*}[ht]
% \centering
% \begin{tabular}{@{}llrrr@{}}
% \toprule
% \textbf{ASR} & \textbf{MT} & \multicolumn{1}{l}{\textbf{BLEU}} & \multicolumn{1}{l}{\textbf{chrF++}} & \multicolumn{1}{l}{\textbf{COMET}}  \\ \midrule
% Whisper-Small  & NLLB-200 600M  & 27.30  & 50.17 & 51.91 \\
% Whisper-Small  & NLLB-200 3.3B  & 27.39  & 49.65 & 52.01 \\
% Whisper-Medium & NLLB-200 3.3B  & 27.45  & 49.64 & 51.74 \\ \bottomrule
% \end{tabular}
% \caption{Performance of Cascaded System Based on BLEU, chrF++, and COMET Scores.}
% \label{tab:perfomance-submissions}
% \end{table*}

% Please add the following required packages to your document preamble:
% \usepackage{booktabs}
\begin{table*}[!htbp]
\centering
\begin{tabular}{@{}llllrrr@{}}
\toprule
\textbf{System} & \textbf{ASR} & \textbf{MT} & \textbf{Type} & \multicolumn{1}{l}{\textbf{BLEU}} & \multicolumn{1}{l}{\textbf{chrF++}} & \multicolumn{1}{l}{\textbf{COMET}} \\ \midrule
\multirow{2}{*}{\textbf{Primary}} & \multirow{2}{*}{Whisper-Medium} & \multirow{2}{*}{NLLB 200 3.3B}  & Baseline  & 0.72  & 14.28 & 16.23 \\
                                & & & Finetuned & \textbf{27.45} & \textbf{49.64} & \textbf{51.74} \\ \midrule
                                
\multirow{2}{*}{\textbf{Contrastive 1}} & \multirow{2}{*}{Whisper-Small}  & \multirow{2}{*}{NLLB 200 3.3B}  & Baseline  & 0.51  & 13.41 & 11.9  \\
                                & & & Finetuned & \textbf{27.39} & \textbf{49.65} & \textbf{52.01} \\ \midrule
                                
\multirow{2}{*}{\textbf{Contrastive 2}} & \multirow{2}{*}{Whisper-Small}  & \multirow{2}{*}{NLLB 200 600M} & Baseline  & 0.41  & 13.21 & 10.69 \\
                                & & & Finetuned & \textbf{27.30}  & \textbf{50.17} & \textbf{51.91} \\ \bottomrule
\end{tabular}
\caption{Performance of the baseline and finetuned cascaded systems based on BLEU, chrF++, and AfriCOMET (COMET) scores. The approaches we followed, including fine-tuning and data augmentation, have considerably improved the quality of Bemba-to-English speech translation. The models were evaluated using the test split of the Big-C dataset.}
\label{tab:eval-submissions}
\end{table*}

\begin{table}[htp]
\centering
\begin{tabular}{llr}
\toprule
\textbf{Model}                  & \textbf{Type} & \multicolumn{1}{l}{\textbf{WER}} \\ \midrule
\multirow{2}{*}{Whisper-Small}  & Baseline             & 157.5                \\
                                & Finetuned            & \underline{35.64}       \\ \midrule
\multirow{2}{*}{Whisper-Medium} & Baseline             & 150.92                \\
                                & Finetuned            & \textbf{36.19}                 \\ \bottomrule
\end{tabular}
\caption{ASR Evaluation: The models were trained with Big-C and BembaSpeech.
%Because "Bemba" is not listed as a supported language in Whisper, the models were trained with "en" as the value for the language parameter.
The performance of the finetuned models outperform the baseline models, indicated by the lower Word Error Rate (WER) scores of the finetuned models compared to the baseline models. The models were evaluated using the test split of the Big-C dataset.}
\label{tab:eval-asr}
\end{table}

\subsection{Data Augmentation}

As explained in Section \ref{sec:synthetic}, we created synthetic data using back-translation to augment our training data \citep{Sennrich2016-BT,Edunov2018-BT,Poncelas2019-BT,Haque2020-Terminology}. Then, we filtered this back-translated data based on generation cross-entropy scores. In our experiments, data augmentation improved the translation quality. As shown in Table~\ref{tab:eval-mt}, when fine-tuning NLLB-200 600M, the models trained with back-translated data outperformed the models trained with only the authentic data.

We tried prepending the back-translated source with the \textit{<bt>} tag, but found removing it achieves better results (cf. Table \ref{tab:eval-bt}). This might be because we filtered the back-translated data, so its quality is good enough that it does not require distinguishing from the authentic data with the \textit{<bt>} tag.

\subsection{Whisper and NLLB-200 Models}

We experimented with both Whisper Small and Whisper Medium to train ASR models. Similarly, we experimented with both NLLB-200 600M and 3.3B to train MT models. For our datasets, the results are comparable (cf. Table \ref{tab:eval-submissions}).

\subsection{End-to-End vs. Cascaded System}

Unlike a cascaded system, an end-to-end speech translation system requires only one model to perform audio-to-text translation \citep{Agarwal2023-IWSLT,Ahmad2024-IWSLT,Moslem2025-SpeechT}. We fine-tuned \mbox{Whisper} directly on the Bemba-to-English Big-C dataset. Table \ref{tab:eval-end-to-end} compares the results of the two systems. Where there is a slight increase in the scores of BLEU And chrF++ of the end-to-end model, the cascaded system outperforms the end-to-end system in terms of the COMET score, while the BLEU score of the end-to-end model is slightly higher.

\begin{table}[htp]
\centering
\resizebox{\columnwidth}{!}{%
\begin{tabular}{llrrr}
\toprule
\textbf{Model}                  & \textbf{Type} & \textbf{BLEU} & \textbf{chrF++} & \textbf{COMET}  \\ \midrule
\multirow{2}{*}{End-to-End}  & Baseline     & 0.09 & 11.85 & 6.9         \\
                                & Finetuned    & \textbf{28.08} & \textbf{49.68} & \underline{48.36}      \\ \midrule
\multirow{2}{*}{Cascaded} & Baseline    & 0.51  & 13.41 & 11.9        \\
                                & Finetuned    & \underline{27.39} & \underline{49.65} & \textbf{52.01}         \\ \bottomrule
\end{tabular}
}
\caption{Comparison of the end-to-end speech translation using Whisper-Small, and the cascaded system that uses Whisper-Small for transcription and then NLLB-200 3.3B for translation. The evaluation uses the test split of the Big-C dataset.}
\label{tab:eval-end-to-end}
\end{table}

\begin{table}[H]
    \centering
    \begin{scriptsize}
    \begin{tabular}{lp{0.85\linewidth}}
    \toprule
      \faVolumeUp & nafwala na amakalashi ku menso \\
      \faClosedCaptioning[regular] & he is wearing glasses as well	 \\
    \faLanguage & He is wearing glasses. \\
      
    \midrule
      \faVolumeUp & Imbwa iyafonka pamoona, ilebutuka palunkoto lwamucibansa \\
      \faClosedCaptioning[regular] & A dog with a wide nose is running on the lawns of the football ground \\
    \faLanguage & A dog with a pointed nose is running on the lawn\\
    
    \midrule
      \faVolumeUp & Akamwanakashi nakemya ukuulu mumuulu ukulwisha ukutoba aka lipulanga. \\
      \faClosedCaptioning[regular] & She has her leg in the air attempting to break a board. \\
    \faLanguage & A child has lifted one leg in an attempt to hit a wood. \\

    \midrule
      \faVolumeUp & Kunuma yabo kuli notu ma motoka tulya ba bonfya mu ncende iya talala nge iyi baliko. \\
      \faClosedCaptioning[regular] & There are also small vehicles that they use in cold places behind them. \\
     \faLanguage & Behind them are vehicles that they use in cold places like this one. \\

    \midrule
      \faVolumeUp & abaume Bali pa mutenge yanganda umo afwele ishati lya mitomito ilyamaboko ayatali elyo me tolishi lya makumbimakumbi \\
      \faClosedCaptioning[regular] & Of the men on the roof of the house, ine is wearing a long sleeved grey shirt and a blue trousers. \\
     \faLanguage & Men are on the roof of the house, one is wearing a grey long sleeved shirt and a blue trousers. \\
     
    \midrule
      \faVolumeUp & Ifi bafwele kunsapato fyakutelelela nga baya mukwangala umu mwine muli ice. \\
      \faClosedCaptioning[regular] & These on their shoes are for sliding when the to play on the ice. \\
     \faLanguage & These shoes they are wearing are for sliding when they are going to play in ice. \\

    \midrule
      \faVolumeUp & Namayo ale enda mumusebo nabika nomwana pamabeya. \\
      \faClosedCaptioning[regular] & A woman is walking in the road with a child on her showders. \\
     \faLanguage & A woman is walking in the road with a child on her shoulder. \\

    % \midrule
    %   \faVolumeUp & Nimbona, iyi incende balateyelako ifyangalo ifyapusana pusana, emulandu wine kuli bambi abaletmba umupila wakuminwe elyo bambi bapoosele amano mukutamba ifyangalo fimbi fye. \\
    %   \faClosedCaptioning[regular] & I have seen it seems there are different games being played that's why others are looking the other way while others are watching volleyball. \\
    %  \faLanguage & I have seen. This place is full of different games that's why others are watching basketball and others are just watching other games. \\

    \midrule
      \faVolumeUp & Namayo naikata ifyakulya pa mbale mukati ke tuuka. \\
      \faClosedCaptioning[regular] & A woman is holding food on a plate inside a shop. \\
     \faLanguage & A woman is holding food on a plate inside a shop. \\

    \midrule
      \faVolumeUp & Nangu limbi kuli bamo abamufulwishe. \\
      \faClosedCaptioning[regular] & Or maybe someone has made him upset. \\
     \faLanguage & Or maybe someone has upset her. \\

    \midrule
      \faVolumeUp & Abantu bane bali umuli ifimabwe ifikulu  nga nshi  kabili nafwala ne fimpopo ku mitwe yabo \\
      \faClosedCaptioning[regular] & four people are inside an area with large rocks and they are wearing helmets \\
     \faLanguage & Four people are in a place full of rocks and they are wearing helmets. \\

    \midrule
      \faVolumeUp & Akamwana kambi balekafuula amasapato kuli kafundisha wakako. \\
      \faClosedCaptioning[regular] & Another child's shoes being taken off by the instructor. \\
     \faLanguage & One of the pupils is being removed the shoes by the teacher. \\

    \midrule
      \faVolumeUp & Afwile alefwaya afike pampela ya lumpili. Pantu icishimbi ekete.Eco babomfya abatemwa ukuniine mpili \\
      \faClosedCaptioning[regular] & Maybe he wants to reach the top of the mountain. The rode metal he is carrying, it is mostly used when one is climbing the mountains. \\
     \faLanguage & Obviously he wants to reach the top of the mountain because this metal he is holding is used by mountain climbers. \\

    % \midrule
    %   \faVolumeUp & kuli umwaume umo umutali uufwele ifyakufwala ifya fita no mwanakashi uufwele ifyakufwala ifya fita uuli no mushishi wa buta na beminina panse ye tuka \\
    %   \faClosedCaptioning[regular] & There is one tall man wearing black clothes and a woman wearing black clothes with white hair standing outside a store. \\
    %  \faLanguage & There is a tall man wearing black clothes and a woman wearing black clothes and has white hair are standing outside a shop. \\
     
    \bottomrule
    \end{tabular}
    \end{scriptsize}
    \caption{Examples of sentences in Bemba, their English translations from the Big-C dataset, and generated translations using Whisper-Medium and NLLB-200 3.3B.}
    \label{tab:examples}
\end{table}

\section*{Acknowledgements}

We would like to thank Kreasof AI for supporting this work through providing the first author with computational resources.

\bibliography{paperpile,citations} 

\end{document}